\documentclass[letterpaper, 10 pt, conference]{ieeeconf}  

\IEEEoverridecommandlockouts                              

\usepackage{relsize}
\usepackage{graphicx}
\usepackage{lipsum}
\usepackage{float}
\usepackage{epsfig} 
\usepackage{xcolor}
\usepackage[bottom]{footmisc}
\usepackage{anyfontsize}
\usepackage[document]{ragged2e}
\usepackage{t1enc}
\usepackage{comment}
\usepackage{multicol}
\usepackage[
    backend=biber,
    style=numeric,
    ]{biblatex}
\usepackage{hyperref}
\hypersetup{
    colorlinks=true,
    linkcolor=blue,
    filecolor=magenta,      
    urlcolor=cyan,
    citecolor=teal,
}
\usepackage{subfig}
\usepackage{subcaption}
\usepackage[font=scriptsize]{caption}
\usepackage{amsfonts, amssymb, amsmath}

\title{\LARGE \bf
Learning-Based Fault Detection for Legged Robots in Remote Dynamic Environments
}

\author{Abriana Stewart-Height$^{1}$, Seema Jahagirdar$^{2}$ and Nikolai Matni$^{3}$
\thanks{$^{1}$Abriana Stewart-Height is a Postdoctoral Fellow with the Department of Mechanical Engineering, Massachusetts Institute of Technology, Cambridge MA 02139, USA
    {\tt\small abrianas@mit.edu}}%
\thanks{$^{2}$Seema Jahagirdar is a MS graduate of the Department of Electrical and Systems Engineering, University of Pennsylvania, Philadelphia, PA, 19104, USA
    {\tt\small seema.jahagirdar1@gmail.com}}%
\thanks{$^{3}$Nikolai Matni is an Associate Professor with the Department of Electrical and Systems Engineering, University of Pennsylvania, Philadelphia, PA, 19104, USA
    {\tt\small nmatni@seas.upenn.edu}}%
}

\begin{document}

\maketitle
\thispagestyle{empty}
\pagestyle{empty}
\justifying
\begin{abstract}
Operations in hazardous environments put humans, animals, and machines at high risk for physically damaging consequences. In contrast to humans and animals, quadruped robots cannot naturally identify and adjust their locomotion to a severely debilitated limb. The ability to detect limb damage and adjust movement to a new physical morphology is the difference between survival and death for humans and animals. The same can be said for quadruped robots autonomously carrying out remote assignments in dynamic, complex settings. This work presents the development and implementation of an off-line learning-based method to detect single limb faults from proprioceptive sensor data in a quadrupedal robot. The aim of the fault detection technique is to provide the correct output for the controller to select the appropriate tripedal gait to use given the robot's current physical morphology.
\end{abstract}

\section{INTRODUCTION}
\noindent Critical field operations (e.g., wildfire suppression, inspection of critical infrastructure, radiation monitoring, disaster response) often require human workers to perform in harsh environmental conditions, putting themselves at risk for serious injury or loss of life. The application of robots to field assignments in extreme environments could reduce the chance of severe harm to human workers. Yet, these increasingly difficult environments also in puts the machines in danger of severe damage. Therefore, it is imperative for their long-term usage that these rescue robots can detect, isolate, and recovery from faults on their own.

It is a challenging endeavor to recognize errors in real-time and to determine an appropriate solution that abates immediate human intervention. However, improvements in resiliency of legged robots in high risk settings would greatly benefit numerous fields related to public safety and security. In prior work, the design of agile fault recovery gaits for a damaged quadruped robot was presented \cite{stewart-height_limb-loss_2024}. Yet, before these alternative behaviors can be used, the robot must first identify the damage. The task of identifying errors in a system can be accomplished through various means \cite{miljkovic_fault_2011}. However, in this work, we focus on legged robots, constraining our attention to research that employs fault detection\footnote{Fault detection is often discussed in partnership with fault diagnosis, so this manuscript uses the terms interchangeably} for robotic systems. 

In robotics, fault detection is designed to identify either unanticipated or anticipated failures. Researchers have proposed different methods for tackling this problem in legged machines based on the type of failure. Model-based optimization techniques have frequently been used to expose unanticipated faults \cite{bongard_resilient_2006, chen_fault-tolerant_2014, cui_fault-tolerant_2022,gor_fault_2018, zhang_novel_2021}, while not as common in the literature, machine learning approaches to unanticipated faults have also been proposed \cite{miguel-blanco_general_2020, sun_comparative_2021}. These methods have shown some promise, but they require a highly accurate model of the robot (or of the robot's leg), which may not always be readily available. In contrast, past research addressing anticipated damage in legged systems has focused heavily on simulations, which predict numerous potential failures for a given robot and generate functional\footnote{Here, functional defined by some performance criteria given by the authors.} gaits for each one. Several authors have proposed evolutionary-based algorithms (EAs) to address anticipated faults \cite{park_automated_2013,chattunyakit_bio-inspired_2019}. EAs often need large search spaces to provide accurate results, making them computationally expensive \cite{park_automated_2013} and time-consuming \cite{zhang_resilient_2017}. The exception to using evolutionary techniques for anticipated fault detection, \cite{kaushik_fast_2021} proposed an algorithm that combines meta-learning and situation embeddings to adapt a dynamical model quickly to a new situation. Simulations allow for researchers to validate their algorithms consistently without damaging the robot or waiting for batteries to charge, however, physical implementation and evaluation is necessary for practical use.

For real world applications, an ideal fault detection approach is computationally inexpensive, performs efficiently (not time-consuming) in real-time (i,e., online), does not require a full dynamical model of the robot (which may not be available), can process large amounts of gait data, and addresses unanticipated faults. Given the needs of the algorithm for the desired task of fault detection, I have chosen to use a deep learning approach to identify limb damage in a quadrupedal robot. Although there are four types of deep learning methods, the two most often found in the literature for applications of fault detection are supervised and unsupervised learning \cite{abid_review_2021}. Unsupervised learning techniques have been shown to be very advantageous for fault diagnosis --- more specifically anomaly detection --- in robotics \cite{khalastchi_fault_2018,marsland_-line_2005,puck_ensemble_2022,narayanan_learning_2018,azzalini_minimally_2021,ren_deep_2019,trivun_resilient_2017,ma_noise-excitation_2023,schnell_robigan_2022} and computer science \cite{chandola_anomaly_2009,omar_machine_2013,nassif_machine_2021}. Anomaly detection refers to the problem of finding patterns in data that do not conform to expected behavior \cite{chandola_anomaly_2009}. Considering the previously discussed literature, it seems that a deep learning methodology --- supervised or unsupervised --- is the most informed choice for our desired task because of its ability to process large amounts of data, automatically execute feature extraction, and end-to-end problem solving capability. The main contribution of this work is the implementation of an offline unsupervised learning technique to accurately identify and locate a single limb failure in a quadruped robot. With this learning algorithm, it is possible to input either a unlabeled dataset of both non-damaged and damaged gait data and predict the faulty limb. Empirical data collected from a physical robot are used to train the deep learning models and test their validity.

The paper is organized as follows. Section \ref{sect:learningtask} defines the problem we are looking to solve and outlines the dataset. Section \ref{sect:method} details the learning approach for detection of limb anomalies and the results of the applying the technique to the generated dataset. Section \ref{sect:results} summarizes the main contributions of our approach and discusses key experimental observations. Section \ref{sect:conclusion} provides concluding remarks.

\section{LEARNING TASK} \label{sect:learningtask}
\begin{figure}[!htb]
    \centering   
    \includegraphics[width=0.45\textwidth]{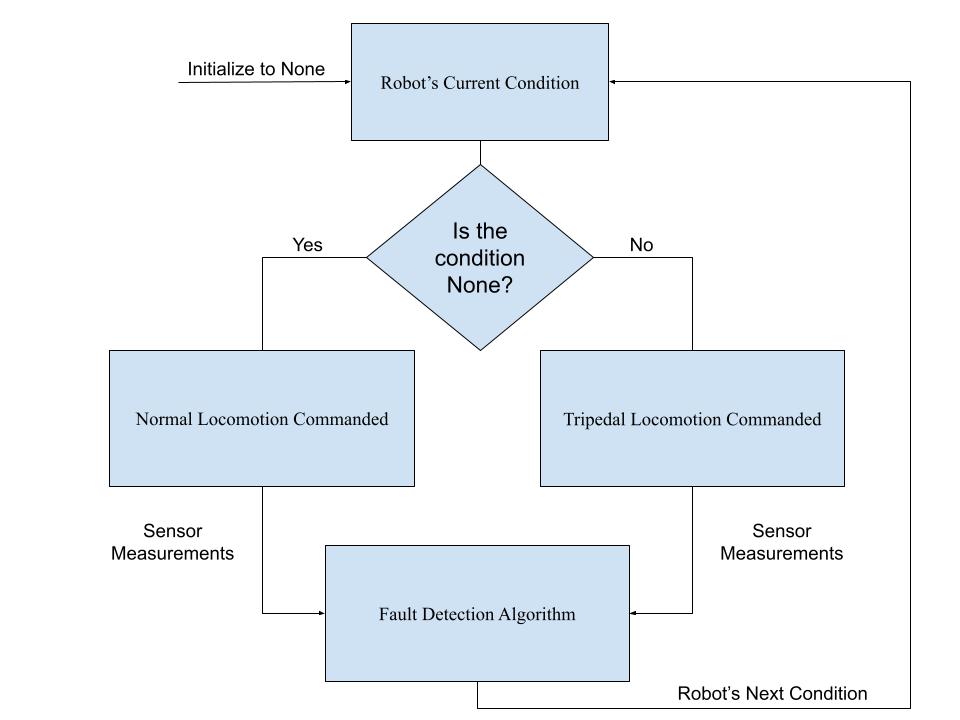} 
    \caption{Flowchart of Proposed Fault Recovery Framework}
    \label{fig:recoveryframework}
\end{figure}

\noindent The novel fault recovery gaits presented in our prior work \cite{stewart-height_limb-loss_2024} assumed a known limb fault. Learning algorithms offer an attractive body of techniques for automating the acquisition of fault detection capabilities that can provide key information about the robot’s condition to the low-level controller. Ideally, the robot should learn to autonomously detect its damaged limb and use the output of the offline learning scheme to determine the correct physical morphology for the desired tripedal dynamic gait in real time. For instance, if the learning algorithm predicts that the left-front (LF) limb is damaged, that output would inform the controller to execute the selected gait (e.g., fore-aft tripedal pronking) under the assumption that the left-front limb is missing. Figure \ref{fig:recoveryframework} gives an overview of the flow between the fault recovery  gaits and the fault detection strategy.

%
%

\subsection{Building the Data Set} \label{sect:data}
\noindent In \cite{stewart-height_limb-loss_2024}, a description of the different experimental setups for fore-aft pronking for both the intact and damaged quadruped robot was provided. The data collected during these experiments as well as during preliminary runs, are combined into a single data set to be used in the training, validation, and testing of the learnt model. The final dataset consists of $170$ data points. Each data point represents one trial of a quadrupedal robot executing a desired fore-aft pronking gait over a distance of either 1 or 2 meters at various distinct fore-aft speeds. In this work, only the gait data of the normal case are considered as input for training the model, but the model is trained on data from both normal and failure cases. The cases refer to the five conditions of the damaged robot which correspond to the location of the damage: none (N), left-front limb (LF), right-front limb (RF), left-back limb (LB), and right-back limb (RB). The data set consists of multiple runs from the five different damage conditions. Each damage condition contains trial runs at different speeds.

Each data point, for each limb of the robot, contains the following sensor measurements: limb number (0-3), time (in ms), limb position (in m), limb velocity (in m/s) --- both reported in body coordinates ----, current for hip motor $\alpha$, and current for hip motor $\beta$ (both in amps). These measurements were acquired by printing the output data to a serial monitor\footnote{CoolTerm is the simple serial port terminal application used for capturing the output data to a text file for each experiment.} at a sampling rate of $60$ Hz and capturing it in a text file.  

Table \ref{table_data_set} shows the number of data points for each condition. The data size of each condition varies for two reasons. First, the data was collected for two different series of experiments (fore-aft pronking with a fixed angled
stepping policy and fore-aft pronking with a Raibert stepping policy, so the total trial time (i.e., the total time spent collecting data during a single trial) --- which includes the robot operating at rest --- varies from 90-120 seconds. Second, some of the data points represent test runs before the experimental procedures were finalized, so they are measured over 1 meter instead of 2 meters. The reason for including non-standardized fore-aft pronking gait data is to have a more practical collection of data to train on. In the real world, the amount of data collected will vary in time and distance traveled. Last, the original individual data samples include sensor measurements for standing and pronking, so data collected while the robot was not in motion is filtered out, which reduces our data set size but allows the model to train on just pronking gait data.

\begin{table}[!ht]
\caption{Dataset Breakdown}
\label{table_data_set}
\begin{center}
\begin{tabular}{|c c c c|}
\hline
Condition & Number of Points & Size & Average Time per Trial\\
\hline
None & 70 & 36 MB & $\sim$90s \\
LF & 30 & 8 MB & $\sim$60s \\
RF & 15 & 8 MB & $\sim$120s \\
LB & 24 & 29 MB & $\sim$90s \\
RB & 31 & 32 MB & $\sim$90s \\
\hline
\end{tabular}
\end{center}
\end{table}

\subsection{Problem Statement}
\noindent In a remote outdoor unstructured environment, the data collected by the robot while performing a gait would not be labeled because the robot is actively collecting and processing the inputs simultaneously. Therefore it is desirable for the task of fault detection that the chosen deep learning algorithm is capable of accurately identify the faulty limb with an unlabeled dataset. In this section, an unsupervised deep learning approach is explored for diagnosing a ``lost'' limb with the dataset described in Section \ref{sect:data}. Now the aim is to train a model to identify patterns that deviate significantly from expected behavior, which is known as anomaly detection \cite{chandola_anomaly_2009}. Anomalies are data points that differ substantially from most of the data, representing unusual events or errors in a dataset. In this case, limb damage is the anomaly I seek to detect. 

\section{METHODS}\label{sect:method}
\noindent For the task of anomaly detection, the proposed approach is the use of an autoencoder's reconstruction loss as a measure of in/out distribution of the data, with the in-distribution being regular behavior, and out of distribution being faulty (i.e., anomalous) behavior. Autoencoders are a type of NN that imposes a bottleneck on input data, outputting a compressed representation of the original input \cite{noauthor_autoencoders_2023}. This compressed representation is often interpreted as the ``essential features'' of the data, which addresses the previously discussed problem of how to extract the ``right'' features from the data points. There are two steps to training an autoencoder model: encoding and decoding. The encoder takes the input data and creates a compact representation. In practice, the encoder/decoder can be any type of NN. In this work, the key feature is the bottleneck of a lower-dimensional representation. 

Our specific implementation of the autoencoder consists of the following layers: the first layer has 8 neurons and uses rectified linear unit (ReLU) activation function, the second layer has 4 neurons with ReLU activation, and the third layer has 2 neurons with ReLU activation. In contrast, the decoder takes the encoded data and attempts to reconstruct the original input data. It consists of the following layers: the first layer has 4 neurons and uses ReLU activation function, the second layer has 8 neurons with ReLU activation, and the third layer has 4 neurons with ReLU activation. The autoencoder model is trained to minimize the difference between the input and the reconstructed output. The loss function chosen is to evaluate the model is the mean squared error (MSE), which is also known as the \textit{reconstruction loss} \cite{noauthor_autoencoders_2023}. While training, the encoder learns to extract important features from the proprioceptive sensor data and the decoder learns to reconstruct the input data using the encoded representation. Reconstruction loss is the difference between the input and reconstructed data. A lower reconstruction loss indicates that the model is accurately learning the underlying patterns of data.

\section{EXPERIMENTAL RESULTS} \label{sect:results} 
\begin{figure}[!htbp]
    \centering    
    \includegraphics[width=0.45\textwidth]{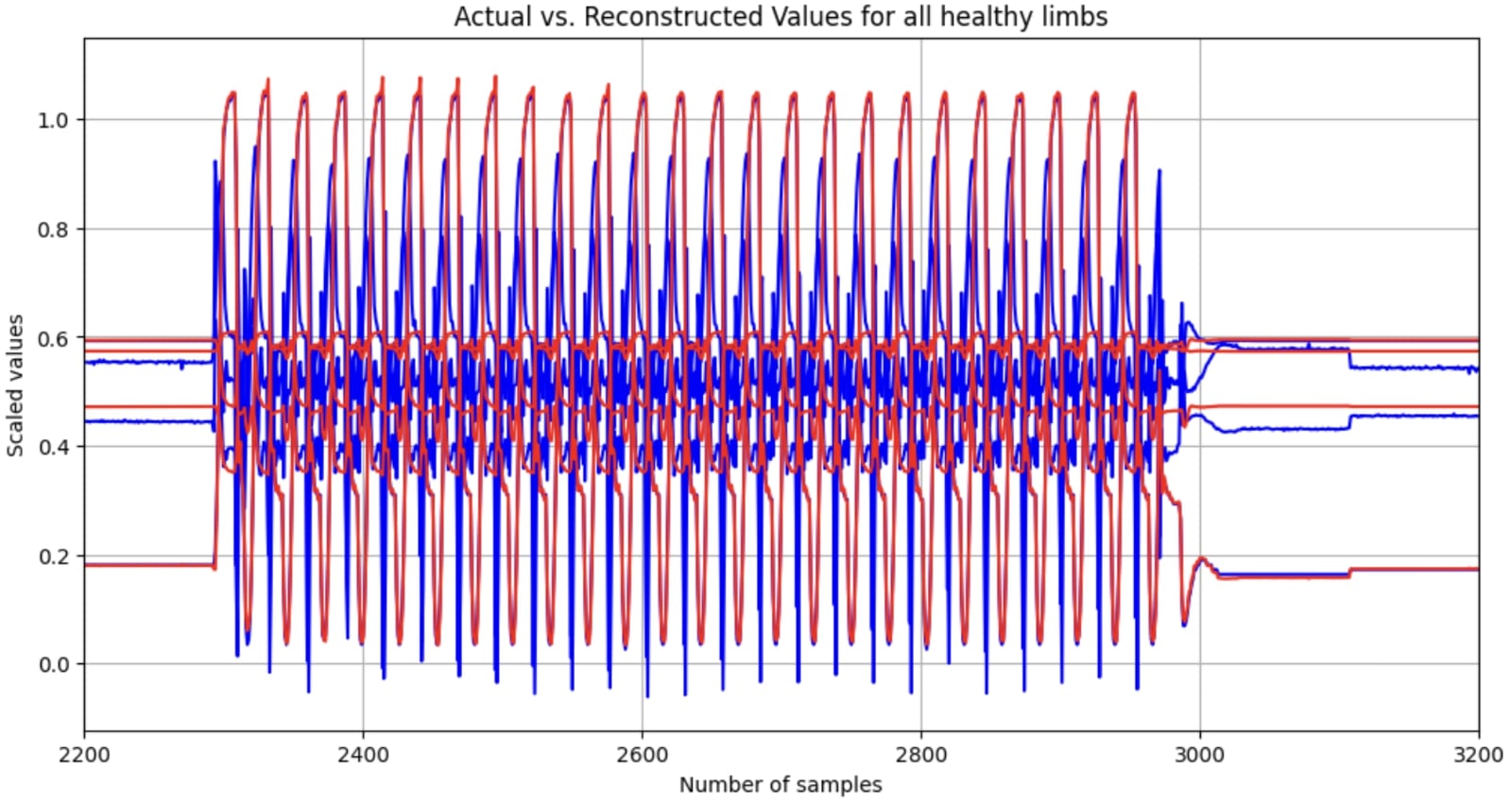} 
    \caption{Actual input values (blue) Vs. Reconstructed input values (red) for the intact quadrupedal robot pronking}
    \label{fig:intact_robot_sensor_values}
\end{figure}

\begin{figure}[!htbp]
    \centering    
    \includegraphics[width=0.45\textwidth]{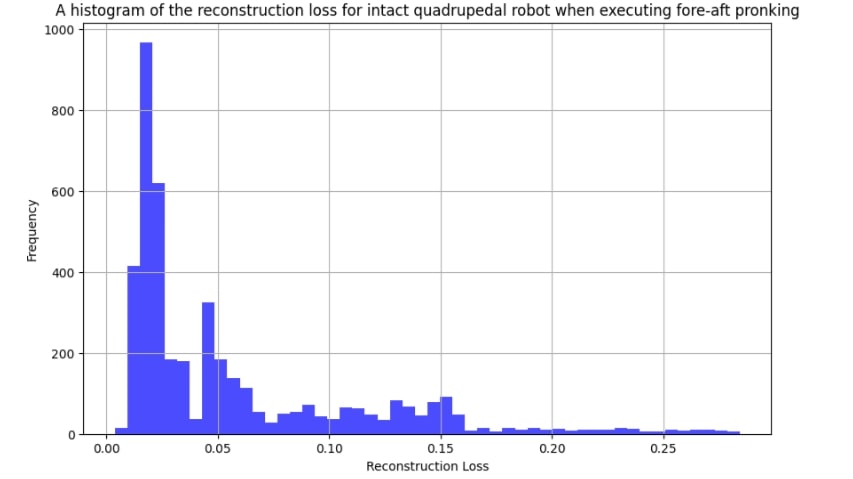} 
    \caption{Distribution of reconstruction errors for the intact quadrupedal robot when executing fore-aft pronking.}
    \label{fig:intact_robot_hist}
\end{figure}

\noindent The data collected from the gait performance of the intact quadrupedal robot is used to train, validate and test the autoencoder model in the ratio 60:20:20, respectively. The model is only trained with data of the no damage condition. The data collected from each limb damage condition is used for validation and testing only. The data is pre-processed using MixMaxScaler function from scikit-learn \cite{scikit_sklearnpreprocessingminmaxscaler_2023} to scale the input features (i.e., sensor measurements) such that their range is between 0 to 1 prior to being fed into the model. After training, the model is evaluated on test data to reconstruct data from both the intact and damaged quadrupedal robot. The threshold for anomaly detection is based on the autoencoder reconstruction errors for normal and faulty data. The threshold is calculated from the mean ($\mu$) and standard deviation ($\sigma$) of the reconstruction errors of the training data. The threshold is calculated as shown in Equation \ref{eqn:threshold} below. A constant multiplier `x' is found by experimenting with different threshold values to determine the best trade-off between false positives and false negatives. For the threshold, it was determined that it is more important (i.e. less consequential) that the anomalous data be correctly identified than non-anomalous data being correctly identified.
\begin{equation} \label{eqn:threshold}
    threshold = \mu_{\text{training loss}} + x*\sigma_{\text{training loss}}
\end{equation}
If the reconstruction loss is higher than the threshold, the sample is flagged as an anomaly. Figure \ref{fig:intact_robot_sensor_values} shows actual sensor values and their corresponding reconstructed values for fore-aft quadrupedal pronking. In the plots for both the original input (red) and reconstructed input (blue), there are four lines each representing the four sensor measurements: limb position, limb velocity, motor $\alpha$ current, motor $\beta$ current. The vertical axis displays the values of the sensor measurements after they have been standardized. Figure \ref{fig:intact_robot_hist} presents a histogram of reconstruction loss for data collected from the intact quadrupedal robot. A histogram shows the frequency of numerical data using rectangles. The height of a rectangle (on the y-axis) represents the distribution frequency of a variable (the amount, or how often that variable appears) \cite{tague_quality_1995}. The reconstruction loss for the normal or healthy data samples is frequently low, which suggests that the model predicts correctly with high accuracy.
\begin{figure}[!htbp]
\centering
\begin{subfloat}
    \raggedright
    \centering
    \includegraphics[width=0.47\textwidth]{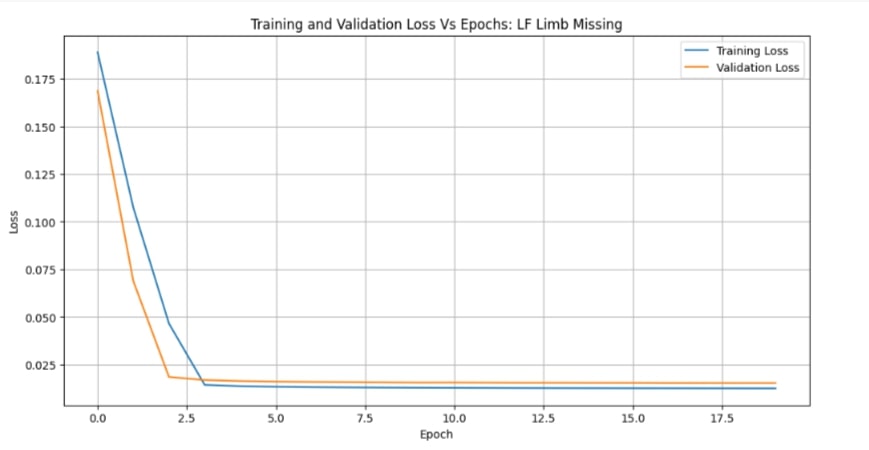}
    \caption*{(a)}
\end{subfloat}
\begin{subfloat}
    \raggedright
    \centering
    \includegraphics[width=0.47\textwidth]{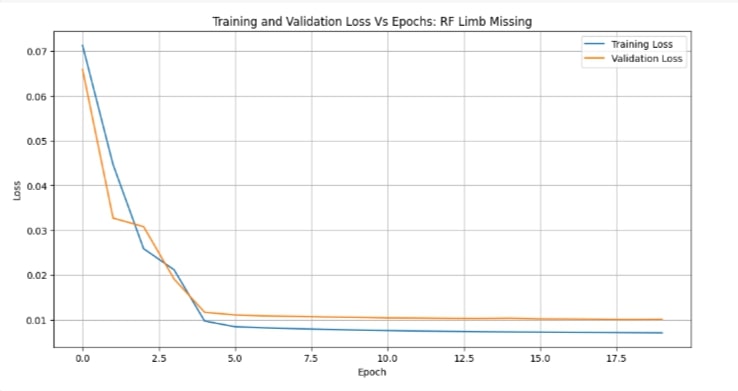}
    \caption*{(b))}
\end{subfloat}
\caption{Training and Validation Loss Vs. Number of Epochs: (a) LF Limb Missing; (b) RF Limb Missing}
\label{fig:loss_epochs}
\end{figure}
\begin{figure}[!htbp]
\centering
\begin{subfloat}
    \raggedright
    \centering
    \includegraphics[width=0.47\textwidth]{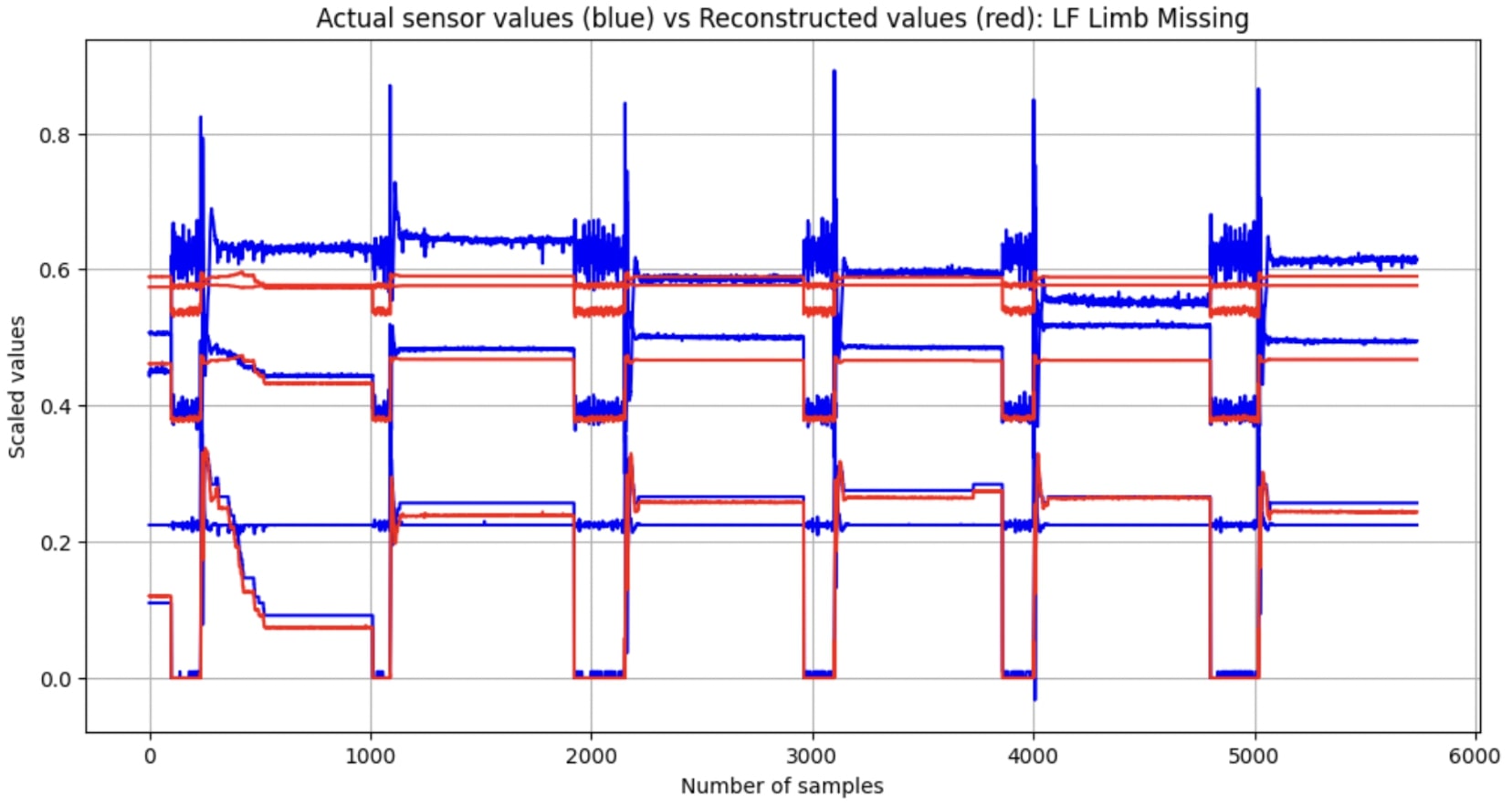}
    \caption*{(a)}
    \label{fig:lf_missing_robot_sensor_values}
\end{subfloat}
\begin{subfloat}
    \raggedright
    \centering
    \includegraphics[width=0.47\textwidth]{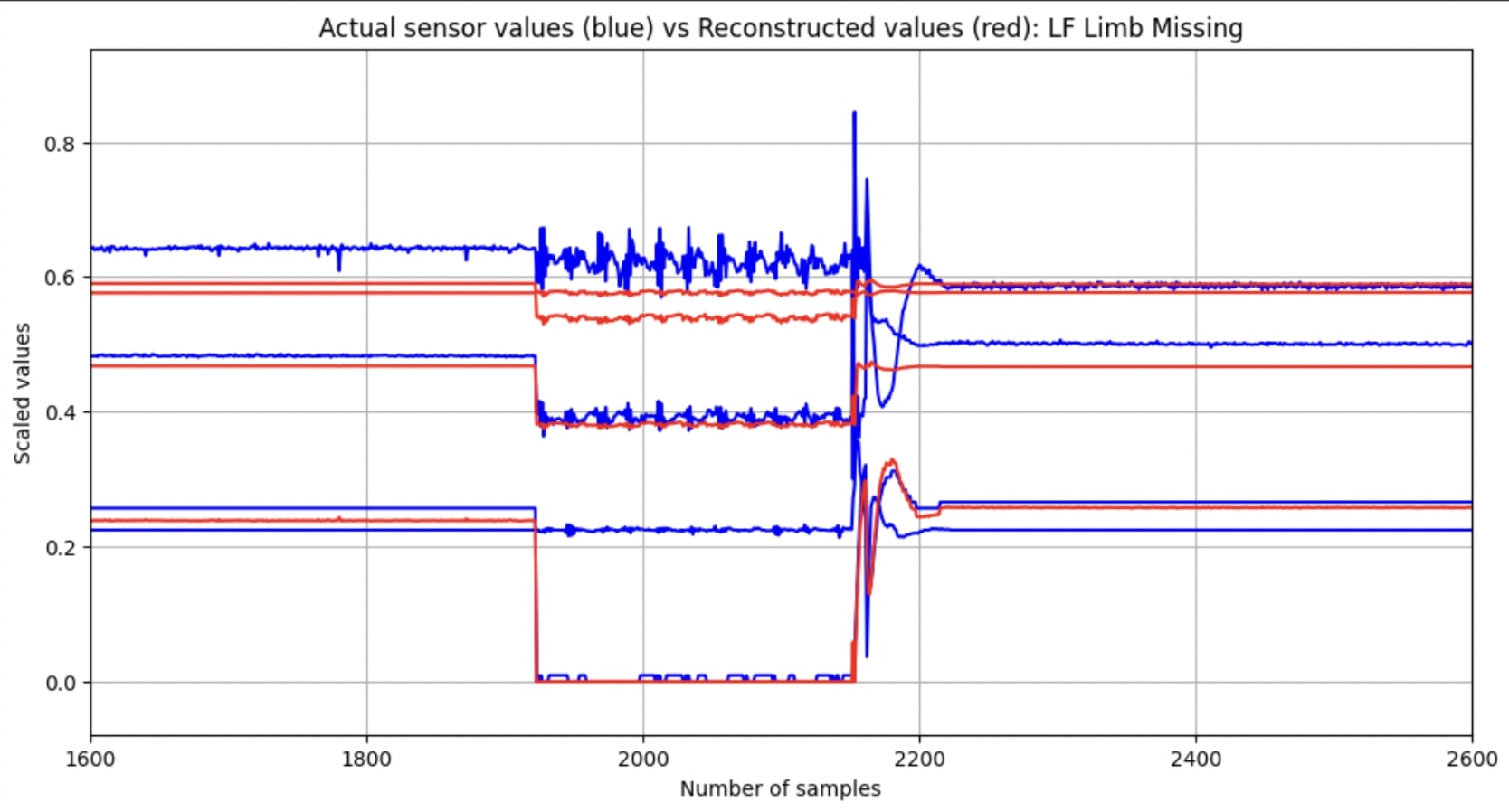}
    \caption*{(b))}
    \label{fig:lf_missing_robot_sensor_values_zoom}
\end{subfloat}
\caption{Actual input values (blue) vs Reconstructed input values (red) for LF Limb Missing: (a) Plot of scaled sensor measurements over the number of samples; (b) A small time window of data from the original plot.}
\label{fig:sensor_values_lf}
\end{figure}
\begin{figure}[!htbp]
\centering
\begin{subfloat}
    \raggedright
    \centering
    \includegraphics[width=0.47\textwidth]{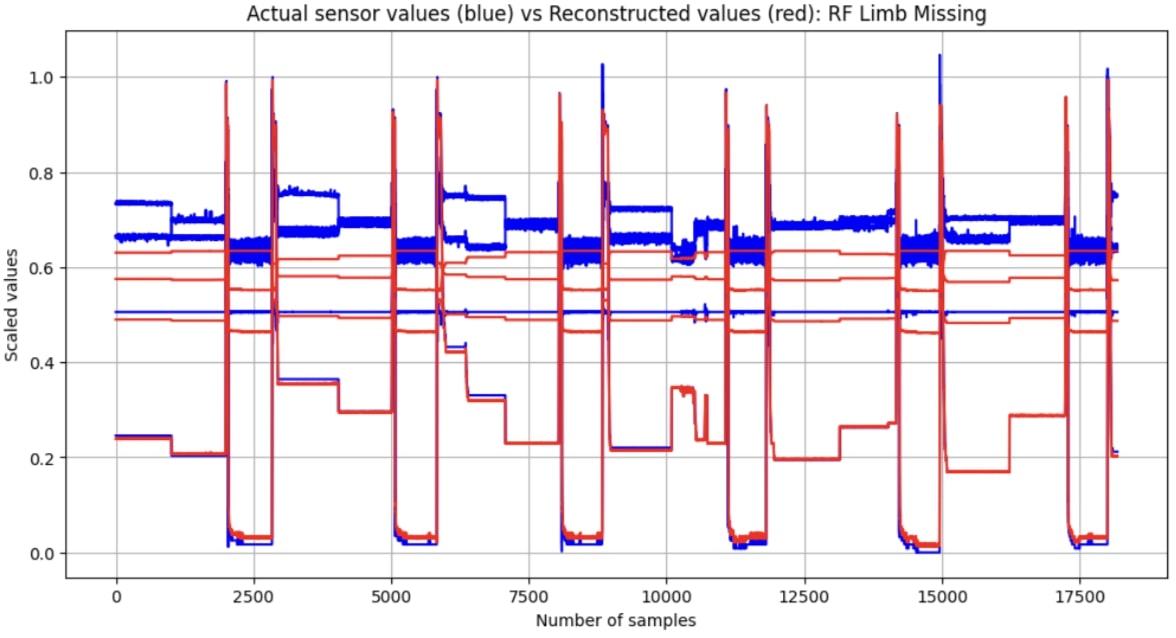}
    \caption*{(a)}
    \label{fig:rf_missing_robot_sensor_values}
\end{subfloat}
\begin{subfloat}
    \raggedright
    \centering
    \includegraphics[width=0.47\textwidth]{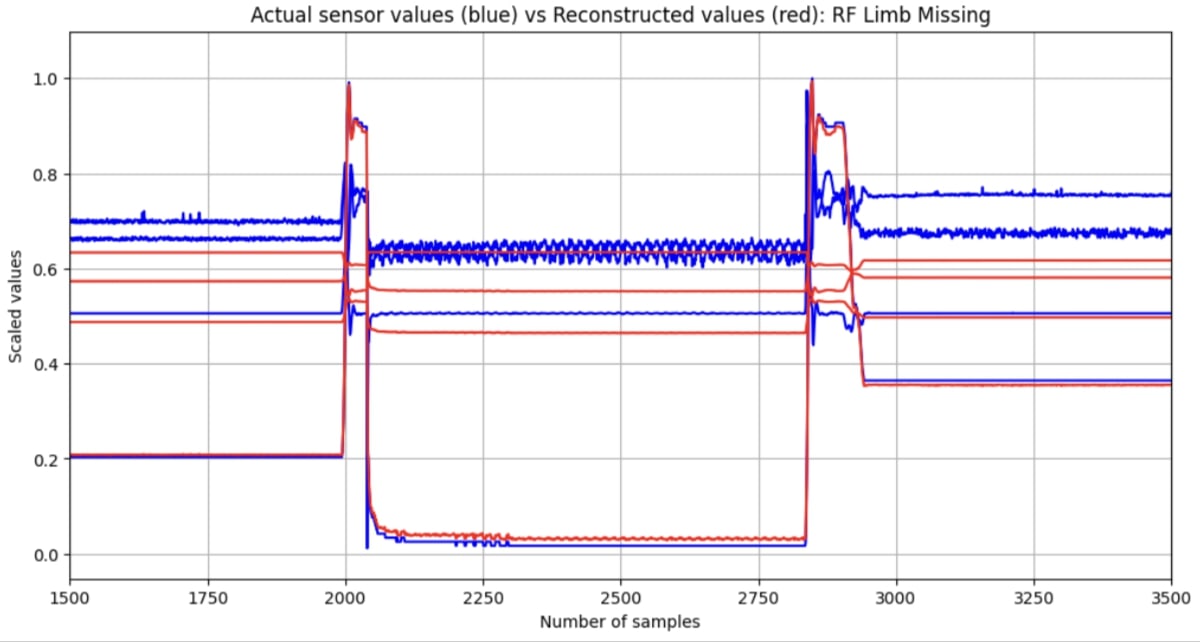}
    \caption*{(b))}
    \label{fig:rf_missing_robot_sensor_values_zoom}
\end{subfloat}
\caption{Actual input values (blue) vs Reconstructed input values (red) for RF Limb Missing: (a) Plot of scaled sensor measurements over the number of samples; (b) A small time window of data from the original plot.}
\label{fig:sensor_values_rf}
\end{figure}
\begin{figure}[!htbp]
\centering
\begin{subfloat}
    \raggedright
    \centering
    \includegraphics[width=0.47\textwidth]{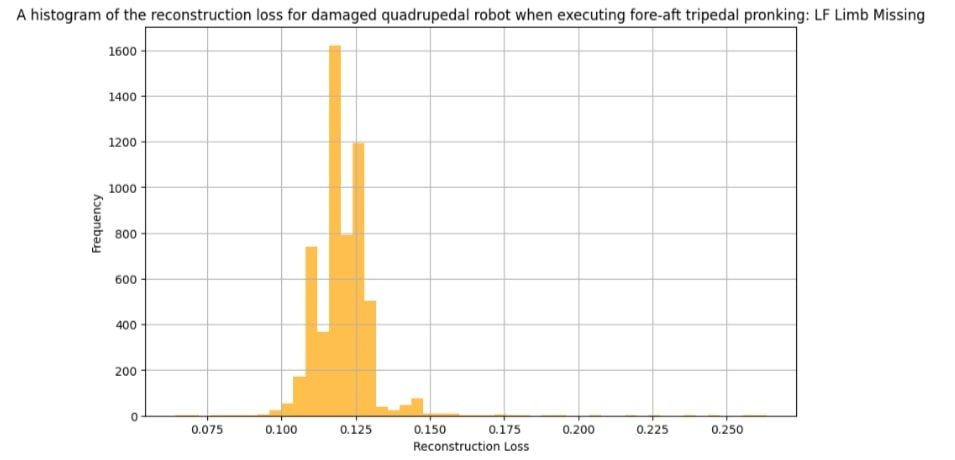}
    \caption*{(a)}
    \label{fig:lf_missing_robot_hist}
\end{subfloat}
\begin{subfloat}
    \raggedright
    \centering
    \includegraphics[width=0.47\textwidth]{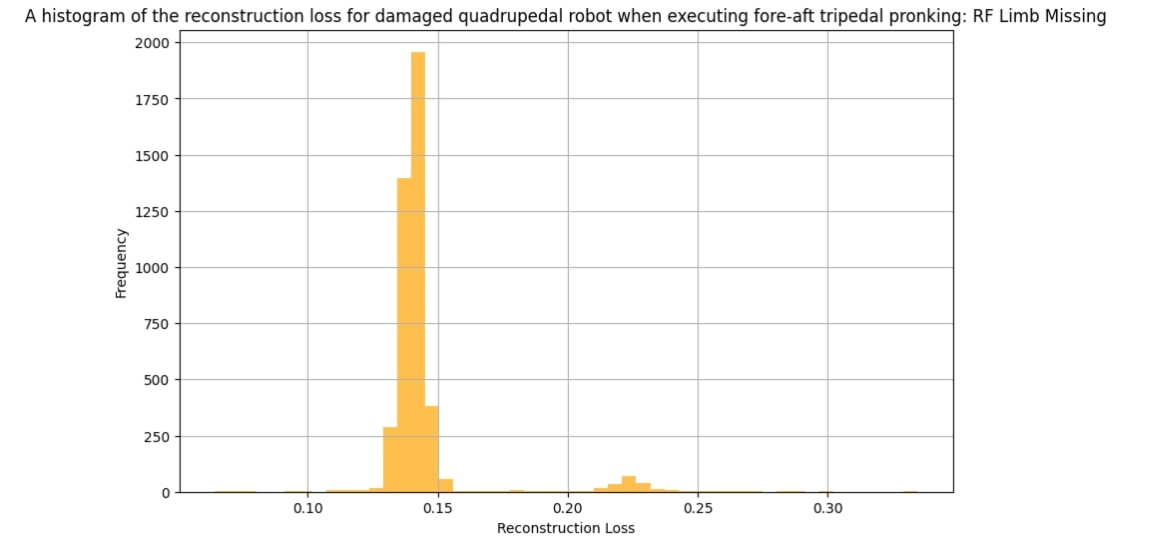}
    \caption*{(b))}
    \label{fig:rf_missing_robot_hist}
\end{subfloat}
\caption{Distribution of reconstruction errors for damaged quadrupedal robot when executing fore-aft tripedal pronking: (a) LF Limb Missing; (b) RF Limb Missing}
\label{fig:tripods_histogram}
\end{figure}
\begin{figure}[!htbp]
\centering
\begin{subfloat}
    \raggedright
    \centering
    \includegraphics[width=0.47\textwidth]{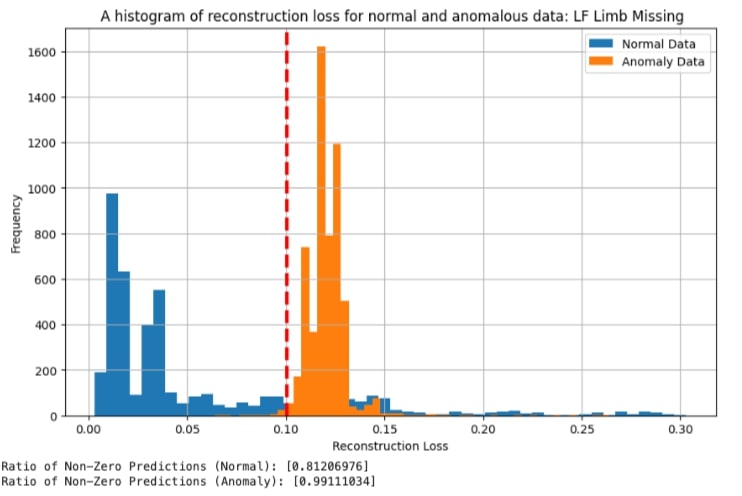}
    \caption*{(a)}
    \label{fig:lf_missing_comp_hist}
\end{subfloat}
\begin{subfloat}
    \raggedright
    \centering
    \includegraphics[width=0.47\textwidth]{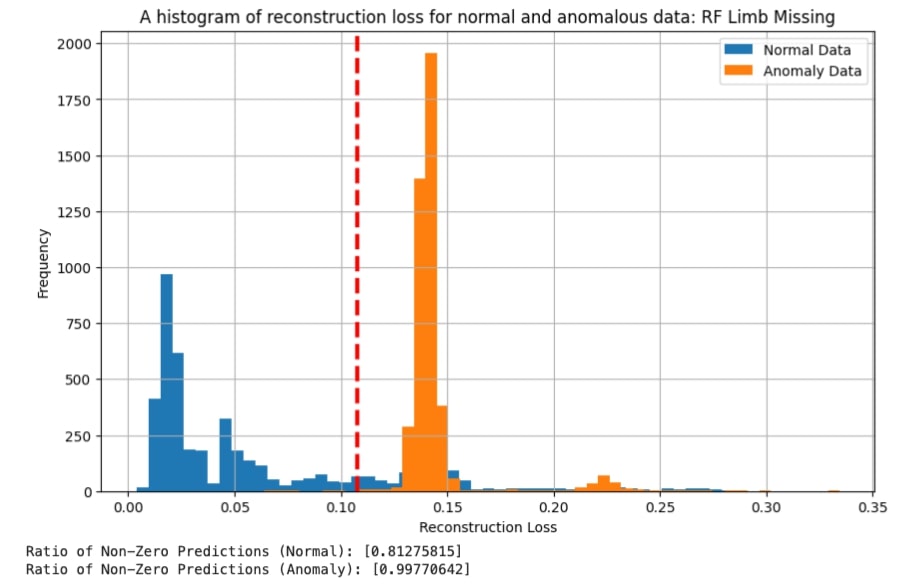}
    \caption*{(b))}
    \label{fig:rf_missing_comp_hist}
\end{subfloat}
\caption{Distribution of reconstruction loss for normal and anomalous data along with the threshold: (a) LF Limb Missing; (b) RF Limb Missing}
\label{fig:compare_robot_hist}
\end{figure}
Figure \ref{fig:loss_epochs} shows the training and validation losses for different numbers of epochs for two different damage conditions. Figures \ref{fig:sensor_values_lf} and \ref{fig:sensor_values_rf} show the actual sensor values and the reconstructed values for the two distinct damage conditions. Figure \ref{fig:tripods_histogram} presents a histogram of reconstruction loss for data collected from a damaged Minitaur in two different tripod configurations. The reconstruction loss for the damaged data samples is not nearly as low as for the non-damaged data samples, suggesting that the model often correctly identifies these samples as anomalies. Figure \ref{fig:compare_robot_hist} displays a histograms of the reconstruction loss for normal (non-damaged) and anomalous (damaged) data along with the threshold for two different damage conditions. The results state for the individual models (LF and RF respectively) that $84.39$\% and $82.48$\% of non-damaged samples are flagged as non-damaged while $99.89$\% and $96.86$\% of damaged samples are flagged as anomalous.

\section{DISCUSSION AND CONCLUSION} \label{sect:conclusion}
\noindent In pursuit of a deep learning method capable of identifying faults with limited time-series data, this paper first explored an unsupervised deep learning approach based on the rich prior literature of using autoencoders for anomaly detection. For practical usage, the ability to train a model on an unlabeled dataset versus a labeled data set is more appealing. These conclusions led us to by pass multi-class classification problem and look into techniques geared towards anomaly detection. Preliminary results confirm that in less than two minutes the featured autoencoder is able to predict a specific limb is damaged when comparing its gait data to the normal (non-damaged) gait data for that particular limb. The challenge now introduced is if we ensemble these individual models into one large model, will the model perform better, worse, or about the same than the individual ones. We hypothesize that the ensembled model will perform better than the individual models, especially when provided with more gait data.

Empowering legged robots with the ability to recover from limb faults while operating remotely in unstructured dynamic environments is a challenging, but rewarding problem to tackle. The potential benefits would greatly assist developers and users in a variety of applications including (but not limited to) space exploration, forestry, environmental monitoring, wildfire suppression, and disaster recovery. This work introduces an offline deep learning technique for limb fault detection in correspondence to previous work which presented a bio-inspired family of stable dynamical gaits. Future work will focus on the integration of these two distinct systems for fault detection and recovery in quadrupedal robots that operate in highly complex, dynamic environments.

\addtolength{\textheight}{-12cm}   



\section*{ACKNOWLEDGMENT}
\noindent The authors would like to thank Dr. Dan E. Koditschek for his helpful discussions bearing upon the preliminary analysis and contributions of this research. This work was supported in part by an NSF Graduate Research Fellowship held by the first author and in part by ONR grant N00014-16-1-2817 held by Dr. Koditschek, sponsored by the Basic Research Office of the Assistant Secretary of Defense for Research and Engineering.

%
\printbibliography

\end{document}